\documentclass[journal,twoside,web]{ieeecolor}
\usepackage{lcsys}
\usepackage{cite}
\usepackage{amsmath,amssymb,amsfonts}
\usepackage{algorithmic}
\usepackage{graphicx}
\usepackage{textcomp}
\usepackage{comment}
\def\BibTeX{{\rm B\kern-.05em{\sc i\kern-.025em b}\kern-.08em
    T\kern-.1667em\lower.7ex\hbox{E}\kern-.125emX}}
    
\usepackage{dsfont}
\usepackage{pdfpages}
\usepackage[bookmarks=false,colorlinks=true,citecolor=blue,linkcolor=blue]{hyperref}
\usepackage{mathtools}
\usepackage{bbm}
\usepackage{epsfig} 
\usepackage{times} 
\usepackage{color}
\usepackage[english]{babel}
\usepackage{subfigure}
\usepackage{nicefrac}
\usepackage{bbm}
\usepackage[T1]{fontenc}

\def\argmin{\operatornamewithlimits{arg\,min}}
\def \interpolates{\cong}

\newcommand{\defeq}{\doteq}

\newcommand{\newpart}{}
\newcommand{\blue}{\color{blue}}

\newcommand{\BP}{\mathbb{P}}

\newcommand{\CC}{\mathcal{C}}

\newcommand{\CO}{\mathcal{O}}

\newcommand{\CH}{\mathcal{H}}

\newcommand\convergealmostsurely{\mathrel{\stackrel{\makebox[0pt]{\mbox{\normalfont\tiny a.s.}}}{\longrightarrow}}}

\makeatletter
\newcommand*{\rom}[1]{\expandafter\@slowromancap\romannumeral #1@}
\makeatother
\def \CG{\mathcal{G}}

\def \CZ{\mathcal{Z}}

\def \RKHS{\mathcal{H}}

\def \one{\mathds{1}}

\newcommand{\tr}{^\mathrm{T}}  
\newcommand{\RR}{\mathbb{R}}

\usepackage{xparse} 

\newcounter{assumptioncounter}
\renewcommand{\theassumptioncounter}{A\arabic{assumptioncounter}}

\NewDocumentEnvironment{assumption}{o}{%
  \refstepcounter{assumptioncounter}%
  \begin{itemize}
    \item[{\theassumptioncounter}]%
    \IfValueT{#1}{\label{#1}}%
}{%
  \end{itemize}
}

\allowdisplaybreaks

\newtheorem{theorem}{Theorem}

\newtheorem{definition}{Definition}

\newcommand{\norm}[1]{\left\lVert#1\right\rVert}    
    
\newlength{\dhatheight}

\begin{document}

\title{\LARGE \bf
Derandomizing Simultaneous Confidence Regions for Band-Limited Functions by Improved Norm Bounds and Majority-Voting Schemes}
\author{\hspace*{-12.5mm}Bal{\'a}zs Csan{\'a}d Cs{\'a}ji 
\and\hspace*{8mm} B\'alint Horv\'ath
\thanks{\newpart This research was supported by the European Union project 
RRF-2.3.1-21-2022-00002 within the framework of the Autonomous Systems National Laboratory; 
and by the TKP2021-NKTA-01 grant of the 
National Research, Development and Innovation Office (NKFIH), Hungary.}%
\thanks{\newpart B.~Cs.~Cs\'aji is with HUN-REN SZTAKI: Institute for Computer Science and Control, Hungarian Research Network; and also with Department of Probability Theory and Statistics, Institute of Mathematics, E\"otv\"os Lor\'and University (ELTE), Budapest, Hungary, {\tt\small csaji@sztaki.hu}}%
\thanks{\newpart B.~Horv\'ath is with HUN-REN SZTAKI: Institute for Computer Science and Control, Hungarian Research Network, Budapest, Hungary; and also with  Institute of Mathematics, Budapest University of Technology and Economics (BME), Hungary, {\tt\small balint.horvath@sztaki.hu}}%
}

\hyphenation{pa-ram-et-ri-za-ti-on}

\maketitle
\thispagestyle{empty}
\pagestyle{empty}

\begin{abstract}
Band-limited functions are fundamental objects that are widely used in systems theory and signal processing.
In this paper we refine a recent nonparametric, nonasymptotic method for constructing simultaneous confidence regions for band-limited functions from noisy input-output measurements, by working in a Paley-Wiener reproducing kernel Hilbert space. Kernel norm bounds are tightened using a uniformly-randomized Hoeffding's inequality for small samples and an empirical Bernstein bound for larger ones. We derive an approximate threshold, based on the sample size and how informative the inputs are, that governs which bound to deploy. Finally, we apply majority voting to aggregate confidence sets from random subsamples, boosting both stability and region size. We prove that even per-input aggregated intervals retain their simultaneous coverage guarantee. These refinements are also validated through numerical experiments.
\end{abstract}


\vspace*{-1mm}
\section{Introduction}
\IEEEPARstart{B}{and-limited}
functions have a principal role in many domains; their applications include, e.g., sampling theory (cf.\ Nyquist-Shannon theorem), aliasing prevention, data compression, noisy-channel coding (cf.\ Shannon–Hartley theorem), multiplexing, interference avoidance, frequency-domain identification, model reduction and networked control systems.

This paper addresses the problem of estimating an {\em unknown} band-limited function from a finite sample of noisy input-output data. This is a standard {\em regression} problem, prevalent across statistics, signal processing, and machine learning, where the primary objective is to identify the {\em regression function}, representing the conditional expectation of the output given an input \cite{cucker2007learning}. 
While numerous regression methods exist, including linear regression \cite{gross2003linear}, neural networks \cite{dreyfus2005neural}, and kernel machines \cite{gyorfi2002distribution}, they typically provide {\em point estimates};
while in many cases, such as robust control or risk management, {\em region estimates} are also needed. These often come in the form of {\em confidence regions} and traditional methods construct them based on the asymptotic distribution of scaled estimation errors, providing only {\em asymptotic} guarantees.
{\em Gaussian Processes} (GPs) represent another approach to region estimation, which yield {\em finite sample} guarantees, but assume that the data is jointly Gaussian \cite{rasmussen2006gaussian}; which may be restrictive in certain applications. Hence, {\em distribution-free} guarantees are desirable. 

Recently, building upon the theory of {\em Reproducing Kernel Hilbert Spaces} (RKHSs), particularly {\em Paley-Wiener} spaces, a novel method for constructing simultaneous confidence regions for band-limited regression functions has been developed \cite{csaji2022nonparametric, csaji2023improving, csaji2023eysm}; and a similar deterministic construction has also been proposed \cite{scharnhorst2022robust}. This {\em nonparametric} approach offers significant advantages: it eliminates the need for parametric model specifications and provides {\em distribution-free} inference with {\em nonasymptotic} guarantees, requiring only mild statistical assumptions to achieve {\em simultaneous} coverage for all inputs.

In this paper, we first provide a brief overview of the Paley-Wiener space based method, and then suggest and experimentally evaluate two types of refinements for the construction:\hspace*{-2mm}
\smallskip
\begin{itemize}
    \item[(1)]
    We improve the {\em upper confidence bounds} on the kernel norm needed for the confidence band construction with a {\em uniformly-randomized} Hoeffding's inequality \cite{ramdas2023randomized} for small samples and an {\em empirical} Bernstein bound \cite{maurer2009empirical} for larger ones. The latter cannot be directly applied, as it depends on
    unobserved data, but we show how to overcome this via a {\em semidefinite} optimization formulation.
    We also derive a {\em threshold} to decide which bound to apply.
    \medskip
    \item[(2)] 
    We investigate {\em majority-voting} schemes \cite{gasparin2024merging} for merging confidence regions derived from random {\em subsamples}. We argue that aggregating the abstract regions is not feasible, hence, the confidence intervals of each query input should be combined separately. We show that even though these intervals have heterogenous sizes, the combination retains their {\em simultaneous} coverage guarantee. We also demonstrate empirically that these aggregated regions exhibit reduced variability and yield tighter confidence bands.
\end{itemize}

\section{Kernels and Band-Limited Functions}

Kernel methods have a wide range of applications, e.g., in machine learning, statistics, system identification and signal processing \cite{berlinet2004reproducing}. Their theory is based on the concept of Reproducing Kernel Hilbert Spaces (RKHSs), defined as
\smallskip
\begin{definition}
{\em Let $\mathbb{X}\neq \emptyset$ and $\mathcal{H}$ be a Hilbert space of $f: \mathbb{X} \to \mathbb{R}$ functions, with inner product $\langle\cdot,\cdot\rangle_\mathcal{H}$. If every Dirac (linear) functional, which evaluates functions at a point, $\delta_z: f \to f(z)$, is continuous for all $z \in \mathbb{X}$ at any given $f \in \mathcal{H}$, then $\mathcal{H}$ is called a Reproducing Kernel Hilbert Space (RKHS)}.
\end{definition}
\smallskip
Every RKHS has a unique {\em kernel}, $k: \mathbb{X} \times \mathbb{X} \to \mathbb{R}$. This function is symmetric and positive definite and it also has the \textit{reproducing property}, namely $\langle k(\cdot,z),f \rangle_\mathcal{H} = f(z),$ for each $z \in \mathbb{X}$ and $f \in \RKHS$. From this property, it follows that for any given $z,s \in \mathbb{X}$, we also have
$k(z,s)=\langle k(\cdot,z),k(\cdot,s) \rangle_\mathcal{H}.$
\smallskip
\begin{definition}
\textit{A Paley-Wiener space with parameter $\eta > 0$
is a subspace of $\mathcal{L}^2 (\mathbb{R}^d)$ in which for each function,
the support of its Fourier transform is included in the hypercube $[-\eta, \eta\hspace{0.3mm}]^d$.
}
\end{definition}
\smallskip
Note that constant $\eta$ is a fixed hyper-parameter. Paley-Wiener spaces are
RKHSs \cite{berlinet2004reproducing} with the following {\em strictly} positive-definite {\em reproducing kernel} function. For all $u, v \in \mathbb{R}^d,$ 
$$
k(u,v) \,\doteq \,  \frac{1}{\pi^{d}} \prod_{j=1}^d \frac{\sin(\eta(u_j-v_j))}{u_j-v_j},
$$
where, for convenience, $\sin(\eta \cdot 0)/0$ is defined to be $\eta$. 
It is important to note that, since a Paley-Wiener (PW) space is a subspace of $\mathcal{L}^2$, it inherits its inner product and its norm. Paley-Wiener spaces consist of band-limited functions, whose frequencies are restricted, thus cannot vary arbitrarily rapidly.

\section{Problem Setting}
We are given $(x_1,y_1), \dots, (x_n,y_n) \in \mathbb{R}^d \times \mathbb{R}$, a sample of input-output pairs with an unknown joint distribution. Let
$$y_k \,=\, f_*(x_k)+\varepsilon_k,$$
for $k \in [\hspace{0.5mm}n\hspace{0.5mm}] \defeq \{1,...,n\},$ where $\{\varepsilon_k\}$ are the {\em measurement noises} on the (unknown) ``true'' \textit{regression function} $f_*.$

Our {\em main goal} is to construct simultaneous, nonparametric {\em confidence bands} for the unknown regression function $f_*$. These confidence bands should have guaranteed (user-chosen) {\em coverage} probabilities for finite (possibly small) sample sizes.

Formally, we are looking for a function $I: \mathbb{R}^d \to \mathbb{R} \times \mathbb{R}$, where $I(x) = (I_1(x), I_2(x))$ specifies the {\em endpoints} of an interval estimate for $f_*(x)$, where $x \in \mathbb{R}^d$. The aim is to construct a function $I$ satisfying the following property:
\begin{equation*}
\nu(I) \,\defeq\, \mathbb{P}\big(\hspace{0.3mm}\forall x \in \mathbb{R}^d: I_1(x) \leq f_*(x) \leq I_2(x)\hspace{0.3mm}\big)\, \geq\, 1-\alpha,
\end{equation*}
where $\alpha \in (0,1)$ is a user-chosen {\em risk} probability. The quantity $\nu(I)$ is called the \textit{reliability} of the confidence band.

Now, we define the property of {\em distributional invariance} \cite{csaji2022nonparametric}.\hspace*{-1mm}
\smallskip
\begin{definition}
{\em An $\mathbb{R}^d$-valued random vector $\varepsilon$ is distributionally invariant w.r.t.\ a compact (topological) group of transformations, $(\CG, \circ)$, where ``$\circ$'' denotes 
composition and each $G \in \CG$ is a function $G:\RR^d \to \RR^d$, if for all $G \in \CG$, random vectors $\varepsilon$ and $G(\varepsilon)$ have the same (joint) distribution.}
\end{definition}
\smallskip
An example of this is an {\em exchangeable} vector, since the group of {\em permutation matrices} leave its distribution unchanged.

Our fundamental assumptions for the construction are (we refer to \cite{csaji2022nonparametric} for a discussion and interpretation of them):
\smallskip
\begin{assumption}
\label{assumption-iid-multivariate} 
{\em The dataset $(x_1, y_1), \dots, (x_n, y_n) \in \mathbb{R}^d \times \mathbb{R}$ is an i.i.d.\ sample of inputs and outputs; and for all $k$: $\mathbb{E}[y^2_k] < \infty$.}
\end{assumption}
\vspace{1mm}

\begin{assumption}
\label{assumption-positive-input-density} 
{\em The probability distribution of the inputs, $\{ x_k \}$, is a priori known, it is absolutely continuous, and its density function, $h_*$, satisfies $h_*(x) > 0$, for all $x \in \mathbb{R}^d.$}
\end{assumption}
\vspace{1mm}

\begin{assumption}
\label{assumption-symmetric-noise-refinement-multivariate} For all index $k \in [\hspace{0.5mm}n\hspace{0.5mm}], \: 
\mathbb{E}\big[\varepsilon_k\big]=0$, and
$\varepsilon_k$ is independent from $x_k$. Moreover,
$\varepsilon \defeq (\varepsilon_1, \dots, \varepsilon_n)\tr$ is distributionally invariant w.r.t.\ 
a (known) compact matrix group $\mathcal{G}$.
\end{assumption}
\vspace{1mm}

\begin{assumption}
\label{assumption-Paley-Wiener-space-multivariate} 
{\em Function $f_*$ is from a Paley-Wiener space and there is a (universal) constant $\varrho > 0$, $\forall\,x \in \mathbb{R}^d:  
f_*^2(x) \leq \varrho \, h_*(x).$ }
\end{assumption}
\smallskip

The distributional invariance required by \ref{assumption-symmetric-noise-refinement-multivariate} is met by i.i.d. variables, ensured by \ref{assumption-iid-multivariate}, which are exchangeable. 
Therefore, the group of permutation matrices is our default choice for $\CG$.

\section{Overview of the Original Construction}
\label{sec:overview}

In this section we briefly overview the original construction of \cite{csaji2022nonparametric, csaji2023improving, csaji2023eysm}. We start by recalling some facts about {\em minimum norm interpolation} in RKHSs.
Suppose that we are given a set of input-output pairs, $\{ (x_k, z_k) \}$, where inputs $\{x_k\}$ are {\em distinct}. The element from $\CH$, which interpolates every $z_k$ at the corresponding input $x_k$ and has the smallest kernel norm,
\begin{equation}
\label{min-norm-interpolant}
\bar{f} \defeq \argmin\!\big\{\|\hspace{0.3mm}f\hspace{0.4mm}\|_{\mathcal{H}} \mid f \in \mathcal{H}\hspace{1mm} \&\hspace{1mm} \forall\hspace{0.3mm} k \in [n]: f(x_k) =\, z_k   \big\}   
\end{equation}
exists, if $\CH$ is an RKHS, and it has the following form:
\vspace{-1mm}
\begin{equation*}
\bar{f}(x) \,= \,\sum\nolimits_{k=1}^n \hat{\alpha}_k k(x,x_k),    
\vspace{-1mm}
\end{equation*}
for all $x \in \mathbb{X}$, where $K_{i,j} \defeq k(x_i,x_j)$ is the Gram matrix, $\hat{\alpha} = K^{-1} z$ with $z \defeq (z_1,...,z_n)\tr$ and $\hat{\alpha} \defeq (\hat{\alpha}_1,...,\hat{\alpha}_n)\tr$, assuming $K$ is invertible (guaranteed for strictly positive definite kernels). From this, we also know that $\|\bar{f}\|_\mathcal{H}^2 = \hat{\alpha}\tr K \hat{\alpha}.$

The main steps of the confidence band construction are:
\begin{itemize}
    \item[(1)] First, by building on an earlier work \cite{csaji2019distribution}, we construct a confidence ellipsoid $\CZ \subseteq \mathbb{R}^{n_0}$ which guarantees to contain $f_*(x_k)$, for $k \in [n_0]$, where $n_0 \leq n$ (that is, for a subset of the {\em observed} inputs), with probability of at least $1 - \beta$, where $\beta \in (0,1)$ is user-chosen.
    \item[(2)] Using $\CZ \subseteq \mathbb{R}^{n_0}$, we calculate a stochastic upper bound, $\tau_{\scriptscriptstyle 0}$, for the norm (square) of the  regression function, for a user-chosen $\alpha \in (0,1)$, i.e., $\BP(\|f_*\|^2_\mathcal{H} \leq \tau_{\scriptscriptstyle 0})\geq 1- \alpha$.
    \item[(3)] For any query input $x_0 \in \mathbb{R}^d$, the construction of the confidence interval for $f_*(x_0)$ is as follows: a candidate $z_0$ value is included in the confidence interval if and only if a $z = (z_1, \dots, z_{n_0})\tr \in \CZ$ exists, such that the norm (square) of the minimum norm interpolant for the dataset $\{(x_k, z_k)\}_{k=1}^{n_0} \cup \{(x_0,z_0)\}$ is less than or equal to $\tau_{\scriptscriptstyle 0}$.
\end{itemize}
The interval endpoints of the procedure of Step (3) can be obtained by solving two convex optimization problems, see \cite{csaji2022nonparametric, csaji2023improving, csaji2023eysm}. The construction guarantees that $\nu(I) \geq 1-\alpha-\beta$.

\section{Improving the Kernel Norm Bounds}
\label{sec:norm}

Constructing a guaranteed upper bound for the kernel norm of $f_*$ is an essential task for building the confidence bands. 

As the norm of Paley-Wiener spaces coincides with the $\mathcal{L}^2$ norm, if there were no measurement noises, we could estimate $\|f_*\|^2_{\CH}$ applying {\em importance sampling} \cite{tokdar2010importance}, since as $n \to \infty$, 
\begin{equation}
\label{noise-free-density-convergence}
\begin{aligned}
\frac{1}{n} \sum_{i=1}^n \frac{f_*^2(x_i)}{h_*(x_i)}\,  \convergealmostsurely &\;\,\mathbb{E} \biggr[ \frac{f_*^2(x_1)}{h_*(x_1)} \biggr]  = \int_{\mathbb{R}^d} \frac{f_*^2(s)}{h_*(s)} h_*(s) \mbox{d}s \\ &= \int_{\mathbb{R}^d} f_*^2(s) \mbox{d}s = \norm{f_*}_2^2 = \norm{f_*}_\CH^2,
\end{aligned}
\end{equation}
by the {\em strong law of large numbers} (SLLN). 

In case of measurement noises, we need a method to build a non-asymptotic confidence ellipsoid for a subset of $\{f_*(x_i)\}$, i.e., for some of the noiseless outputs, but only at the {\em observed} inputs.
That is,
for any $\beta \in (0,1)$, we need an ellipsoid $\mathcal{Z}$ with
\begin{equation}
\label{KGP-ellipsoid}
\mathbb{P}\big(\, (f_*(x_1), \dots, f_*(x_{n_0}))\tr \in \mathcal{Z}\,\big) \, \geq \, 1-\beta,
\vspace{0.5mm}
\end{equation}
for some (user-chosen) $ n_0 \in [n]$.
The {\em Kernel Gradient Perturbation} (KGP) algorithm is an example of such methods \cite{csaji2019distribution}. 

Given ellipsoid $\mathcal{Z} \subseteq \RR^{n_0}$, it was shown earlier \cite{csaji2022nonparametric} that
\begin{equation}
\label{norm-estimation-new}
\tau_{\scriptscriptstyle 0} \, \defeq\, \xi^* + \rho \hspace{0.3mm} \sqrt{\ln (1/\alpha)/(2n_0)}.
\end{equation}
serves as a stochastic {\em upper bound} for $\norm{f_*}_{\CH}^2$ and guarantees
\begin{equation*}
\mathbb{P}\big(\norm{f_*}_{\CH}^2 \leq \tau_{\scriptscriptstyle 0} \hspace{0.3mm}\big)\,  \geq \,1-\alpha-\beta,
\end{equation*}
under \ref{assumption-iid-multivariate},  \ref{assumption-positive-input-density}, \ref{assumption-symmetric-noise-refinement-multivariate}, \ref{assumption-Paley-Wiener-space-multivariate}, where $\alpha, \beta \in (0,1)$ are user-chosen risk probabilities, and $\xi^*$ is the {\em optimal value} of the 
problem
\begin{equation}
\label{noisy-norm-max-rewritten}
\mbox{maximize} \; \frac{1}{n_0}\, \sum_{k=1}^{n_0} \frac{{z_k}^2}{h_*(x_k)}\qquad
\mbox{subject to} \; z \in \mathcal{Z}
\vspace{0.5mm}
\end{equation}
which is not convex, but using duality and Schur complements, an equivalent convex semidefinite problem can be built \cite{csaji2023improving,csaji2023eysm}.

The additional term next to $\xi^*$ in \eqref{norm-estimation-new} comes from Hoeffding's inequality \cite{csaji2022nonparametric}. In this section, we describe a uniformly-randomized version of this theorem, which will provide a strictly better upper bound for the kernel norm, while having the same statistical guarantees \cite{ramdas2023randomized}. Later, we also suggest another bound, based on the empirical Bernstein's inequality \cite{maurer2009empirical}, which yields an even  tighter bound, for {\em large samples}.

Now, let us recall the definition of sub-Gaussianity:
\smallskip
\begin{definition}{\em A random variable $X$ is called $\sigma$-sub-Gaussian with variance proxy $\sigma^2$, if for all $\lambda \in \RR$, we have\vspace{-0.5mm}
\begin{equation*}
\mathbb{E}\big[\hspace{0.3mm}\exp(\lambda(X-\mathbb{E}[X]))\hspace{0.3mm}\big] \leq \exp(\lambda^2 \sigma^2/2).
\vspace{1.5mm}
\end{equation*}}
\end{definition}
Hoeffding's inequality can be improved by randomization \cite{ramdas2023randomized}.\hspace{-1mm}
\smallskip
\begin{theorem}[Uniformly-Randomized Hoeffding's Inequality] 
\label{unifran_Hoeffding}
{\em Let $X_1, \dots, X_n$ be i.i.d.\ $\sigma$-sub-Gaussian random variables and $U$ be a 
variable with uniform distribution on $(0,1)$, independent from $X_1, \dots, X_n$. Then, for any $\alpha \in (0,1)$,
\begin{equation*}
\mathbb{P}\Bigg(\mathbb{E}[\bar{X}] \geq \bar{X} + \underbrace{\sigma \sqrt{\frac{2 \ln(1/\alpha)}{n}}+\sigma \frac{\ln(U)}{\sqrt{2n \ln(1/\alpha)}}}_{\textstyle \phi(\sigma,\alpha,n, U)}\Bigg)\hspace{-0.3mm}\leq \alpha,
\vspace{-0.5mm}
\end{equation*}
where $\bar{X}=(1/n)\sum_{i=1}^nX_i$ denotes the sample average.}
\end{theorem}
\smallskip
The original Hoeffding inequality can be obtained for $U=1$ and since $\ln(U) < 0$, the result above provides a tighter bound than the original. Since \ref{assumption-Paley-Wiener-space-multivariate} ensures $f^2_*(x)/h_*(x) \in [\hspace{0.3mm}0,\varrho\hspace{0.3mm}]$, and random variables taking values in $[\hspace{0.3mm}a,b\hspace{0.3mm}]$ are $(b-a)/2$ sub-Gaussian, we can use Theorem \ref{unifran_Hoeffding} to get another upper bound
\begin{equation}
\label{norm-estimation-new-u}
\tau_{\scriptstyle\mathtt{u}} \, \defeq\, \xi^* + \phi(\varrho/2,\alpha,n_0, U),
\vspace{0.5mm}
\end{equation}
which still ensures $\mathbb{P}\big(\norm{f_*}_{\CH}^2 \leq \tau_{\scriptstyle\mathtt{u}} \hspace{0.3mm}\big)\,  \geq \,1-\alpha-\beta$, moreover, it is {\em stricly} better than the original bound, i.e., $\mathbb{P}(\tau_{\scriptstyle\mathtt{u}} < \tau) = 1$.

Now, we suggest an alternative bound, based on an empirical Bernstein's inequality \cite{maurer2009empirical}, which could be used for {\em large samples}, as it has a better convergence rate. 
Before stating the theorem, 
we recall a way to define empirical variance:
\smallskip
\begin{definition} {\em For a sample $X=(X_1, \dots, X_n)$ containing independent random variables, the empirical variance is 
\vspace{-1mm}
\begin{equation*}
V_n(X) \defeq \frac{1}{n(n-1)} \sum_{1 \leq i \leq j \leq n} (X_i-X_j)^2.
\vspace{0.5mm}
\end{equation*}}
\end{definition}
\smallskip
\begin{theorem}[Empirical Bernstein’s Inequality]
\label{thm:emp_Bernstein}
{\em Let $X=(X_1, \dots, X_n)$ be a sample of independent random variables with values in $[\hspace{0.3mm}0, \kappa\hspace{0.3mm}]$. Then, for any $\alpha \in (0,1)$, we have
\begin{equation*}
\mathbb{P}\Bigg(\mathbb{E}[\bar{X}] \geq \bar{X} + \underbrace{\kappa\hspace{0.3mm} \bigg( \sqrt{\frac{2\hspace{0.3mm}V_n(X) \ln (2/\alpha)}{n}} + \frac{7 \ln 2/\alpha}{3(n-1)} \bigg)}_{\textstyle \psi(\kappa, \alpha, n, V_n(X) )}\Bigg)\hspace{-0.3mm}\leq \alpha.
\vspace{2mm}
\end{equation*}}
\end{theorem}
This result also takes the (empirical) variance into account and 
it can provide tighter upper bounds for {\em larger sample sizes}, as the 
$\psi(\kappa, \alpha, n, V_n(X))$ term decreases faster than the 
$\phi(\sigma, \alpha, n)$ term of the Hoeffding type inequality. On the other hand, the extra term added to the sample mean is data-dependent. Hence, even though the $\{f^2_*(x_i)/h_*(x_i)\}$ variables are bounded by $\kappa = \varrho$, in general, they cannot be directly observed. In the special case of no measurement noises $(\varepsilon_k\equiv 0)$, we can use
\begin{equation}
\label{norm-estimation-new-b}
\tau_{\scriptstyle\mathtt{b}} \, \defeq\, \xi^* + \psi(\varrho, \alpha, n_0, V_n(T(y)))
\end{equation}
where $T(y) \doteq (y^2_1/h_*(x_1), \dots, y^2_{n_0}/h_*(x_{n_0}))$ which in this case yields $T(y)= (f^2_*(x_1)/h_*(x_1), \dots, f^2_*(x_{n_0})/h_*(x_{n_0}))$.

In case measurement noises are present, we can exploit ellipsoid $\CZ$. The first idea would be to maximize the whole bound $M(T(z)) + \psi(\varrho, \alpha, n_0, V_n(T(z))$ over $\CZ$, where $M(v)\doteq 1/n_0\sum_iv_i$ is the mean of vector $v \in \RR^{n_0}$. This, however, leads to a challenging optimization problem. Since we only need an upper bound, we can maximize the two terms separately. Maximizing $\psi(\varrho, \alpha, n_0, V_n(T(z)))$ is equivalent to
\begin{equation}
\label{noisy-max-variance}
\mbox{maximize} \; V_n(T(z))\qquad \mbox{subject to} \; z \in \mathcal{Z}
\end{equation}
 where $\mathcal{Z}$ is a confidence ellipsoid satisfying \eqref{KGP-ellipsoid}. The optimal solution of problem \eqref{noisy-max-variance}
is denoted by $z^* \in \CZ$. 
Note that \eqref{noisy-max-variance} is {\em not convex}.
However, it is still a single constraint quadratic optimization problem, which can be reformulated using Lagrange duality theory and exploiting Schur complements as an equivalent 
{\em convex} {\em semidefinite} problem \cite[Appendix B.1]{boyd2004convex}.

Finally, 
$\tau_{\scriptstyle\mathtt{e}} \, \defeq\, \xi^* + \psi(\varrho, \alpha, n_0, T(z^*))$
provides the upper bound
which guarantees that $\mathbb{P}\big(\norm{f_*}_{\CH}^2 \leq \tau_{\scriptstyle\mathtt{e}} \hspace{0.3mm}\big)\,  \geq \,1-\alpha-\beta$.

A natural question that arises is: when should the Bernstein-type bound be preferred over the Hoeffding-type bounds? Exploring this question in its full generality would go beyond the scope of this paper (as it involves data dependent random terms), but as a preliminary analysis, we compare the (standard) Hoeffding bound with the Bernstein bound above, assuming $V_n(X) = \sigma^2$, the true variance, which is a reasonable approximation as $V_n(X)\xrightarrow{a.s.} \sigma^2$ as $n\to \infty$.

First, we can observe that the requirement
\vspace{-0.5mm}
\begin{gather}
\exists\hspace{0.3mm} N: \forall\hspace{0.3mm} n\geq N: \phi(\varrho/2,\alpha,n,1) \leq \psi(\varrho, \alpha, n, \sigma^2)\notag\\[1mm]
\Longleftrightarrow \;\sigma < \sqrt{(\ln\bigl(1/\alpha\bigr))/(4\,\ln\bigl(2/\alpha\bigr))}\label{eq:sigma}
\end{gather}
as we can simplify by $\varrho$ and the second $\CO(1/n)$ term of $\psi$ is asymptotically insignificant compared to the $\CO(1/ \sqrt{n})$ term.

Assuming inequality \eqref{eq:sigma} holds, we can compute an explicit formula for $N$ in terms of $\alpha$ and $\sigma$. After rearrangements and using the solution formula for quadratic equations, we get
\begin{align*}
N &= \Big\lceil ( \varsigma + \sqrt{\varsigma^2 + 4})^2/4 \Big\rceil,\quad\text{with}\\
\varsigma &= \frac{7 \sqrt{2} \ln(2/\alpha)}{3 \left( \sqrt{\ln(1/\alpha)} - 2 \sigma \sqrt{\ln(2/\alpha)} \right)}. 
\end{align*}
We can get rid of the $\sigma$ dependence 
by using that $\sigma \leq \varrho/2$, where  $\varrho$ 
quantifies how informative the inputs are, see \ref{assumption-Paley-Wiener-space-multivariate}.
Then $N = N(\alpha, \sigma) \leq N' = N(\alpha, \varrho/2)$ providing us an (approximate) {\em threshold} for switching to a Bernstein bound.

\begin{figure}[!t]
    \centering
	\hspace*{-2mm}	 	
    \vspace{-1mm}
	\includegraphics[width = \columnwidth]{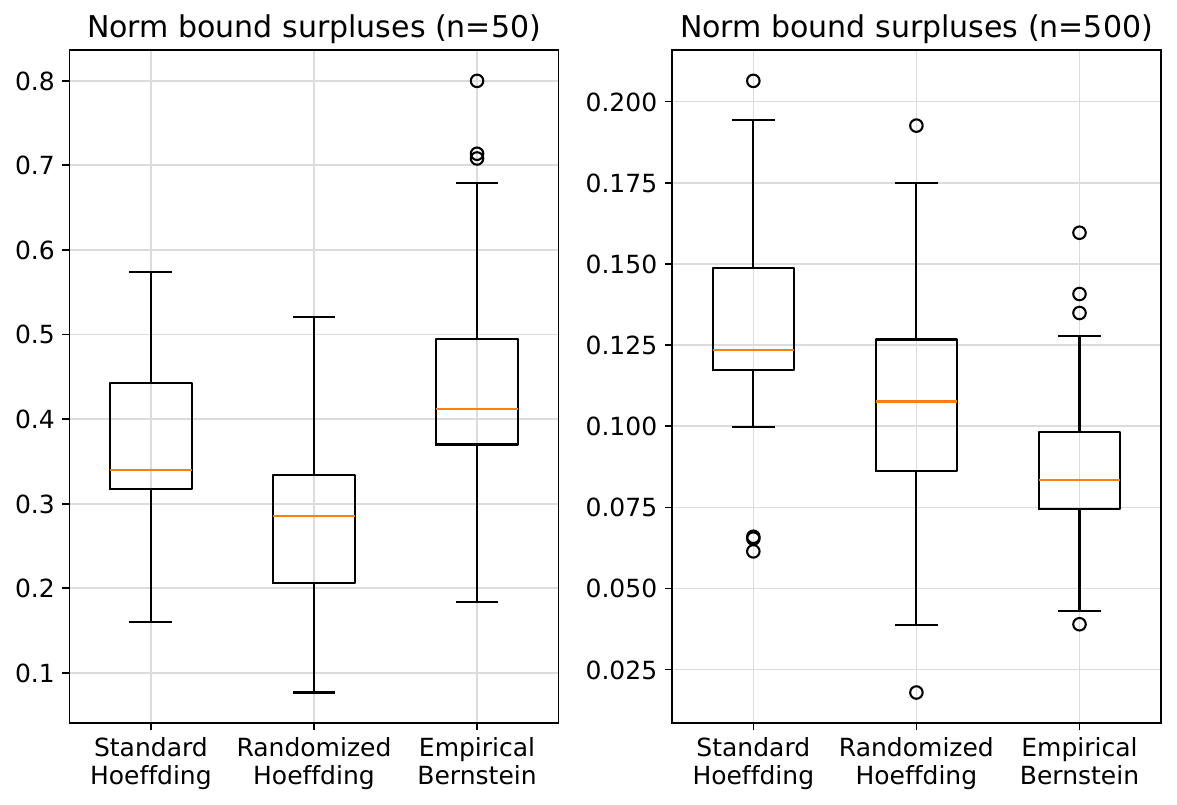} 	
    \caption{Differences of the 
    upper bounds $(\alpha=0.1)$ and the original kernel norm of the regression function for $n=50$ and $n=500$.}
\label{fig:norm-estimation-1}
\vspace*{-1mm}
\end{figure}

To evaluate the two proposed kernel norm upper bounds, we conducted numerical experiments using randomly generated band-limited functions (see Section \ref{sec:numexp}). To isolate the effects of the improved concentration inequalities, we investigated a noise-free setting ($\beta = 0$), where only the inputs, drawn from a Laplace distribution, were random. We computed the norm estimates from \eqref{norm-estimation-new}, \eqref{norm-estimation-new-u}, and \eqref{norm-estimation-new-b} across various $n$ values, assuming a confidence level of $1 - \alpha$ with $\alpha = 0.1$. Figure \ref{fig:norm-estimation-1}
presents box plots from 100 Monte Carlo trials, illustrating the differences between the upper bounds and true norms. The orange line marks the median; boxes span the interquartile range, and whiskers cover values within 1.5 times that range.
The results show that the randomized Hoeffding bound yields tighter estimates for small samples (left part), while the empirical Bernstein bound performs better for larger samples (right part), in harmony with our theoretical analysis.\!\!

\section{Merging Confidence Bands from Subsamples}
\label{sec:Merging-Uncertainty-Sets}

In this section we explore another approach to improve our Paley-Wiener (PW) kernel-based confidence band construction by employing various {\em majority-voting} schemes \cite{gasparin2024merging} to aggregate confidence regions derived from random {\em subsamples}.

The selection of the $n_0 \leq n$ points in the original method, see Section \ref{sec:overview}, is arbitrary due to the i.i.d.\ nature of the data. Therefore, for a fixed $n_0$, we can obtain $\binom{n}{n_0}$ different confidence sets by permuting the sample. Alternatively, we can randomly generate $K$ permutations to obtain $K$ subsamples leading to $K$ confidence regions. We investigate how to combine these sets to achieve an enhanced confidence band.

We now  overview some of the majority voting constructions of
\cite{gasparin2024merging}.
In general, our objective is to build confidence sets that contain a (constant) unknown object $\theta_* \in \Theta$
with a guaranteed probability, e.g., $\theta_* = f_*$ and $\Theta = \CH$. Assume that we are given $K$ different sets (one from each ``agent''), where each set has the same confidence level $\gamma \in (0,1)$, that is
\begin{equation}
\label{uncertainty-sets-property}
\forall\,k \in [K]:\mathbb{P}(\theta_* \in C_k) \geq 1-\gamma.
\end{equation}
Given threshold $t \in [0,1]$ and weight vector $w$, let us define 
\begin{equation*}
\CC(w, t) \defeq  \Bigg\{\, f \in \CH: \sum_{k=1}^K w_k \: \mathbb{I} \big( f \in C_k \big) > t\, \Bigg\},
\end{equation*}
where $w \defeq (w_1, \dots, w_K) \in [\hspace{0.3mm}0,1\hspace{0.3mm}]^K$ with $w \tr \one = 1$ are given weights and $t \in [\hspace{0.5mm}0,1)$.  The set of weights, where each weight is equal will be denoted with $\bar{w} \defeq (1/K, \dots, 1/K) \in [\hspace{0.5mm}0,1\hspace{0.3mm}]^K$.

The first method combines confidence sets through voting: a point is included if over half of the agents vouch for it, i.e.,
\begin{equation}
\label{mv-set}
C^M \defeq \CC(\bar{w}, 1/2).
\end{equation}

Construction \eqref{mv-set} can be generalized for arbitrary thresholds, $\tau \in [\hspace{0.5mm}0,1\hspace{0.3mm})$, that is $C^{\tau} \defeq \CC(\bar{w}, \tau).$ Obviously, $C^M = C^{\tau}$ for $\tau=1/2$. Its guarantees can be summarized as \cite{gasparin2024merging}:
\smallskip
\begin{theorem}
\label{mv-set-tau-theorem-1}
{\em Let $C_1, \dots, C_K$ be $K \geq 2$ different confidence regions satisfying \eqref{uncertainty-sets-property}
and let $\tau \in [\hspace{0.5mm}0,1\hspace{0.3mm})$ be a threshold. Then,
\begin{equation*}
\mathbb{P}(\theta_* \in C^{\tau})\, \geq\, 1-\frac{\gamma}{1-\tau}.    
\end{equation*}
Let $\lambda(C^{\tau})$ be the Lebesgue measure 
of the set $C^{\tau}$. Then, 
\begin{equation*}
\lambda(C^{\tau})\, \leq\, \frac{1}{K \tau} \sum_{k=1}^K \lambda(C_k),
\end{equation*}
for all $\tau \in [\hspace{0.5mm}0,1\hspace{0.3mm})$. Furthermore, if the input regions are $K$ one-dimensional confidence intervals, then for all $\tau \in [\hspace{0.5mm}1/2,1\hspace{0.3mm})$,
\vspace{-0.5mm}
\begin{equation*}
\lambda(C^{\tau})\, \leq\, \max_{k} \lambda(C_k).
\vspace{0.5mm}
\end{equation*}}
\end{theorem}
\smallskip

By processing the confidence sets in a random order, the previous construction can be improved. If $\pi: \{1,2,\dots,K\} \to \{1,2, \dots, K\}$ is a uniformly random permutation, which is independent of the $K$ confidence sets, then 
we can define
\begin{equation}
\label{set-permutation}
C^{\pi} \defeq \bigcap_{k=1}^K C^M(\pi(1) : \pi(k))
\end{equation}
where $C^M(\pi(1) : \pi(k))$ means that we construct a majority vote set from confidence sets $C_{\pi(1)}, \dots, C_{\pi(k)}$.
Region $C^{\pi}$ has the same coverage guarantee, moreover, it is never worse than the majority vote based confidence region, that is \cite{gasparin2024merging}:
\smallskip
\begin{theorem}{\em Let $C_1, \dots, C_K$ be $K \geq 2$ different confidence sets with $1-\gamma$ coverage. If $\pi$ is a uniformly random permutation, which is independent from them, then $C^{\pi}$ is an uncertainty set with $1-2\gamma$ confidence level and $C^{\pi} \subseteq C^M$.}
\end{theorem}
\smallskip
The majority vote can also be improved with the following randomized construction \cite{gasparin2024merging}: let $U$ be a random variable with uniform distribution on $[\hspace{0.3mm}0,1\hspace{0.3mm}]$.
Then, let us define
\begin{equation}
\label{set-thresholding-1}
C^R \defeq \CC (\bar{w}, (1+U)/2).
\end{equation}
This idea has another small variant:
\begin{equation}
\label{set-thresholding-2}
C^U \defeq \CC(\bar{w}, U).
\end{equation}
\begin{theorem}
\label{sets-thresholding-theorem}
{\em Let $C_1, \dots, C_K$ be $K \geq 2$ different confidence sets with $1-\gamma$ coverage. Then, $C^R$ has $1-2\gamma$ coverage and $C^R \subseteq C^M$, while $C^U$ has $1-\gamma$ coverage and $C^R \subseteq C^U$.}
\end{theorem}

If there is any priori information available about the confidence regions, 
the set of weights can be modified to assign bigger values to more ''informative'' sets. If $U$ has uniform distribution on the interval $[\hspace{0.3mm}0,1\hspace{0.3mm}]$, then the weighted set is
\begin{equation}
\label{set-weighting}
C^W \defeq \CC(w, (1+U)/2), 
\end{equation}
where $w \in [\hspace{0.3mm}0,1\hspace{0.3mm}]^K$ satisfies $w \tr \one = 1$. Its guarantees are \cite{gasparin2024merging}:\hspace*{-1mm}
\smallskip
\begin{theorem}
{\em Let $C_1, \dots, C_K$ be $K \geq 2$ different confidence sets 
satisfying \eqref{uncertainty-sets-property}. Then, the weighted randomized set $C^W$ defined by \eqref{set-weighting} gives a level $1-2\gamma$ confidence set. 
Moreover, if $\lambda(C^W)$ is the Lebesgue measure of $C^W$, then
\vspace{-1mm}
\begin{equation*}
\lambda(C^W) \,\leq\, 2 \sum_{k=1}^K w_k \lambda(C_k).
\vspace{1.5mm}
\end{equation*}}
\end{theorem}

In our case, let 
$\pi_1, \pi_2, \dots, \pi_K: [\hspace{0.5mm}n\hspace{0.5mm}] \to [\hspace{0.5mm}n\hspace{0.5mm}]$ 
be random permutations with 
(discrete) 
uniform distribution. Let $\tilde{C}_k$ be the Paley-Wiener kernel-based confidence region obtained by using
sample
${\blue \{(x_{\pi_k(j)}, y_{\pi_k(j)})\}}$, 
for $k \in [K]$. 
Hence, with the choice of $\gamma = \alpha + \beta$, $\mathbb{P}(f_* \in \tilde{C}_k) \geq 1-\gamma$, for $k \in [K]$.
In the abstract sense, these confidence regions are defined as
\begin{align*}
\tilde{C}_k \!\defeq\! \{ f \in \CH \mid\: & \exists\, z \in \mathcal{Z}_k\! :\! f\hspace{-0.3mm} \interpolates\hspace{-0.3mm}\{(x_{\pi_k(j)}, z_{\pi_k(j)})\}\hspace{-0.3mm} \land\hspace{-0.3mm}  \norm{f}_\mathcal{H}^2 \leq \tau_k \},
\end{align*}
where ``$f \interpolates \{(x_i, z_i)\}$'' denotes that $f$ interpolates $\{(x_i, z_i)\}$,
$\tau_k$ is a norm bound with $\mathbb{P}\big(\norm{f_*}_{\CH}^2 \leq \tau_k \hspace{0.3mm}\big) \geq 1-\gamma$, and 
$\mathcal{Z}_k$ is a confidence ellipsoid, see \eqref{KGP-ellipsoid}, based on subsample $k$.

Our first observation is that directly combining these abstract sets using \eqref{set-permutation}, \eqref{set-thresholding-1} and \eqref{set-thresholding-2}, i.e.,
to make a vote for each $f \in \CH$ is practically infeasible. An alternative choice is to 
combine the confidence intervals for each query input, i.e.
\vspace{-0.5mm}
\begin{equation*}
\CC(w,t,x_0)\! \defeq\! \bigg\{\hspace{0.1mm} y \in \mathbb{R}:\!  \sum_{k=1}^K w_k\hspace{0.3mm} \mathbb{I} \big(y\hspace{-0.5mm} \in \hspace{-0.5mm}[I^{(k)}_1\hspace{-0.3mm}(x_0), I_2^{(k)}\hspace{-0.3mm}(x_0)] \big)\hspace{-0.3mm} >\hspace{-0.3mm} t\hspace{0.1mm} \bigg\}\hspace{-0.3mm},
\vspace{-0.5mm}
\end{equation*}
for all $x_0 \in \mathbb{R}^d$, where $I^{(k)}_1(x_0)$ and $I^{(k)}_2(x_0)$ are the endpoints of the confidence interval at input $x_0$ based on subsample $k$. This results in a strictly larger, but practically feasible (easy to compute) set that can be constructed for any query input.

Our next observation is that the confidence intervals obtained by majority voting still satisfy the {\em simultaneous} coverage guarantee. This is far from obvious, because the interval widths are heterogeneous: they vary with the query point, so a confidence interval obtained from a subsample can belong to the majority at some query points but fall outside it at others.

In order to see this, let us introduce the notation
\begin{equation*}
\tilde{D}_k \defeq \big\{ f \in \CH \mid \forall x\in \RR^d: I_1^{(k)}(x) \leq f(x) \leq I_2^{(k)}(x) \big\},
\end{equation*}
where the the interval endpoints satisfy, by construction, that
\begin{equation*}
I_1^{(k)}(x) =\! \min_{f \in \tilde{C}_k} f(x),\quad \text{and}\quad I_2^{(k)}(x) =\! \max_{f \in \tilde{C}_k} f(x),
\end{equation*}
for all $x\in \RR^d$. From this, we immediately have $\tilde{C}_k \subseteq \tilde{D}_k$.

Thus, $\forall\hspace{0.3mm}f \in \CH,\forall\hspace{0.3mm} k \in [K]: \mathbb{I}(f \in \tilde{C}_k) \leq \mathbb{I}(f \in \tilde{D}_k)$, hence
\vspace{-1mm}
\begin{equation}
\sum_{k=1}^K w_k\hspace{0.3mm} \mathbb{I}(f \in \tilde{C}_k) \leq \sum_{k=1}^K w_k\hspace{0.3mm} \mathbb{I}(f \in \tilde{D}_k),
\label{eq:majorityw}
\vspace{-0.5mm}
\end{equation}
where $\{w_k\}$ are nonnegative weights that sum up to one. Let us denote the regions obtained by (potentially weighted) majority voting of $\{\tilde{C}_k\}$ and $\{\tilde{D}_k\}$ sets by $\tilde{C}^M$ and $\tilde{D}^M$, respectively. Then, using \eqref{eq:majorityw}, we (a.s.) have $\tilde{C}^M \subseteq \tilde{D}^M$, thus $\mathbb{P}(f_* \in \tilde{C}^M) \leq \mathbb{P}(f_* \in \tilde{D}^M)$. Therefore, if for all $k$, $\mathbb{P}(f_* \in \tilde{C}_k) \geq 1-\gamma$, we have $\mathbb{P}(f_* \in \tilde{D}^M) \geq 1-2\hspace{0.3mm}\gamma$. Similarly, the 
guarantees of the other constructions also carry over to the aggregation of $\{\tilde{D} _k\}$, meaning that the combined intervals have {\em simultaneous} coverage guarantee, as well.

\begin{figure}[!t]
    \centering
	\hspace*{-2mm}	 	
    \vspace{-1mm}
	\includegraphics[width = \columnwidth]{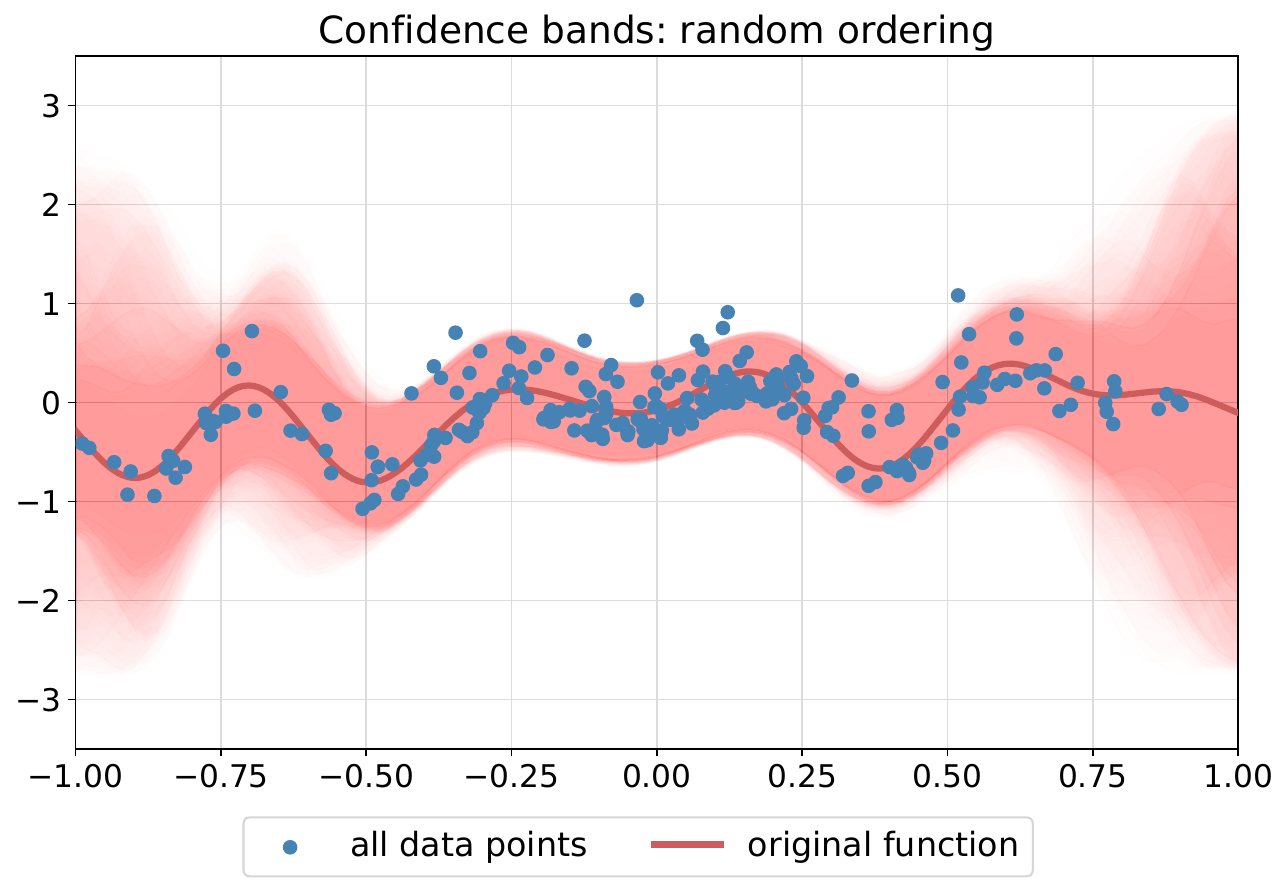} 	
    \caption{$90\,\%$ combined confidence bands using random ordering, \eqref{set-permutation}.}
\label{fig:random-ordering}
\end{figure}
\begin{figure}[!t]
    \centering
	\hspace*{-2mm}	 	
    \vspace{-1mm}
	\includegraphics[width = \columnwidth]{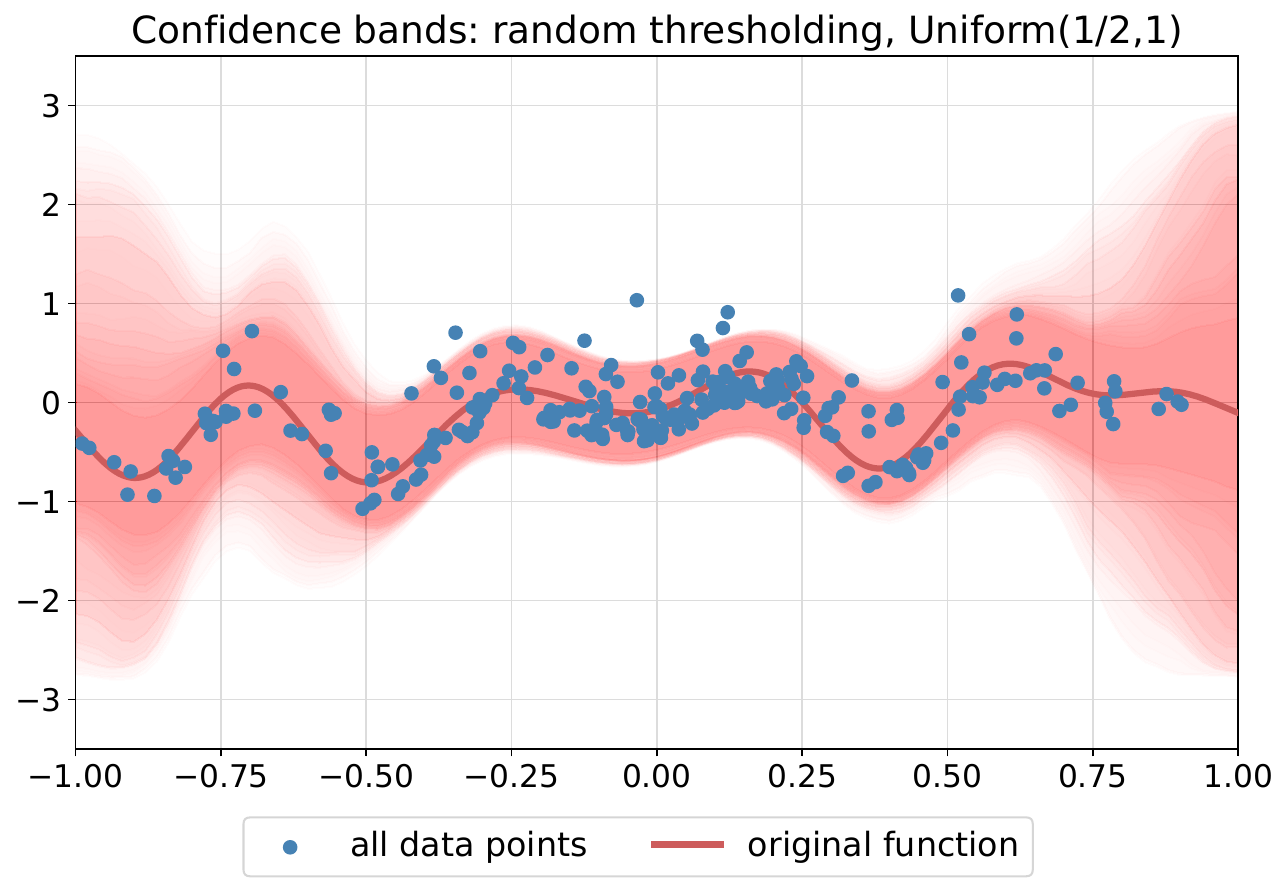} 	
    \caption{$90\,\%$ combined conf.\ bands using random thresholding, \eqref{set-thresholding-1}.}
\label{fig:random-thresholding-1}
\vspace*{-3mm}
\end{figure}

\section{Numerical Experiments}
\label{sec:numexp}

For the numerical experiments, we have constructed the ``true'' band-limited functions, 
$f_*$,
as follows:
first, $20$ random knot points $\{\bar{x}_k\}_{k=1}^{20}$ were generated, with i.i.d.\ uniform distribution on $[\hspace{0.3mm}a,b\hspace{0.5mm}]$. Then, $f_*(x) = \sum_{k=1}^{20} w_k k(x, \bar{x}_k)$ where $\{w_k\}$ had i.i.d.\ uniform distribution on $[\hspace{0.3mm}-1,1\hspace{0.3mm}]$. Finally, the function was normalized, in case its maximum value exceeded $1$. For the observations, inputs $\{x_k\}$ had Laplace distribution with location $\mu=0$ and scale $\zeta>0$. In the norm bound experiments we used $[\hspace{0.5mm}a,b\hspace{0.5mm}]=[\hspace{0.5mm}0, 1\hspace{0.3mm}]$, $\zeta = 1$, and the parameter for the Paley-Wiener kernel was $\eta = 100$. For the majority voting, $[\hspace{0.5mm}a,b\hspace{0.5mm}]=[\hspace{0.3mm}-1, 1\hspace{0.3mm}]$, $\zeta = 0.5$, and $\eta \in \{20, 30\}$, see later.

The experiments regarding norm bounds were presented in Section \ref{sec:norm}, now we present the results regarding the majority voting. We used $n=300$ observations and
the distribution of the $\{ \varepsilon_k \} $ noise terms was $\varepsilon \sim \mbox{exp}(\lambda_0) - \lambda_0$ for $\lambda_0 = 0.3$. These noises are not symmetric, but they still satisfy
\ref{assumption-symmetric-noise-refinement-multivariate}. We created $K=51$ different confidence sets with $n_0 = 17$. Then, $100$ different $\pi: [K] \to [K]$ permutations were used for the Random Ordering (RO) given by \eqref{set-permutation}, while $100$ different $u$ random realizations were taken from the variable $U$, which had uniform distribution on the interval $[\hspace{0.3mm}0,1\hspace{0.3mm}]$ for the two Random Thresholding (RT) methods provided by \eqref{set-thresholding-1} and \eqref{set-thresholding-2}. 

Figures \ref{fig:random-ordering} and \ref{fig:random-thresholding-1}
show the {\em empirical distribution} of the confidence bands given by \eqref{set-permutation} and 
\eqref{set-thresholding-2}, 
respectively.

All combined confidence bands have $1-\gamma$ confidence level with parameter $\gamma = 0.1$; we used $\eta = 20$.
Since the confidence regions in Figures \ref{fig:random-ordering} and \ref{fig:random-thresholding-1} had a $0.9$ confidence level, we 
could apply parameters $\alpha = \beta = 0.025$ in \eqref{KGP-ellipsoid} and \eqref{norm-estimation-new}. 

In these experiments, the observations $\{(x_i,y_i)\}$ were fixed, only the random elements of the voting, that is the random thresholds and permutations changed. Construction \eqref{set-thresholding-2} was quite noisy, while 
\eqref{set-permutation} and \eqref{set-thresholding-1} provided similar results, RO being slightly better than RT with uniform$(0.5,1)$ threshold.

We also performed {\em quantitative} experiments. We used a kernel with $\eta = 30$, calculated
the average, the median and the standard deviation of the diameters for the combined confidence bands using the RO and RT schemes, as follows:
\begin{enumerate}
    \item First, we randomly generated the ``true'' band-limited $f_*$ and the corresponding dataset of observations.
    \item We selected (query) input $x_0$ randomly, according to the known distribution of the inputs (in this case: Laplace). 
    \item We obtained $K = 101$ different confidence intervals for $f_*(x_0)$ by randomly choosing $n_0 \leq n$ points from the sample, applying the original ``standard`` (ST) method.
    \item Then, we calculated the combined intervals based on the $K=101$ confidence intervals, using the (equally weighted) majority-voting schemes \eqref{set-permutation}, \eqref{set-thresholding-1} and \eqref{set-thresholding-2}.
    \item We determined the diameters of each obtained confidence interval, that is the quantity $|\hspace{0.3mm}I_1(x_0) - I_2(x_0)|$, if the endpoints of the interval were $I_1(x_0)$ and $I_2(x_0)$.
    \item We repeated this procedure $100$ times (starting from the construction of $f_*$) and computed the average, the median and the standard deviation of the interval diameters.
\end{enumerate}

The results are summarized in Table \ref{table-diameters}.
It can be observed that the intervals obtained with 
RO and 
RT (with the threshold uniformly chosen between $0.5$ and $1$) have significantly smaller standard deviations in all cases, while the average and median diameter values become closer for larger samples. Still, the majority-voting schemes RO and RT$(0.5,1)$ consistently provided intervals with smaller diameters. Hence, the majority voting approach not only decreases the variability of the results (derandomization), but also provides smaller regions.

{\renewcommand{\arraystretch}{1.5}
\begin{table}[!t]
\vspace{-0.5mm}
\caption{Averages, medians and standard deviations of the diameters for random query inputs, for various constructions.}
\centering
\begin{tabular}{|c|c|c||c|c|c|c|}
\cline{1-7}
Diameter & $n$ & $n_0$ & ST & RO & RT(0.5,1) & RT(0,1)\\ \hline\hline
avg & 100 & 20 & 0.7864 & 0.3689 & 0.3562 & 0.6878\\ \hline
med & 100 & 20 & 0.5717 & 0.0684 & 0.0861 & 0.4086\\ \hline
std & 100 & 20 & 0.7825 & 0.5223 & 0.4911 & 0.7114\\ \hline\hline
avg & 250 & 50 & 0.1911 & 0.1050 & 0.1035 & 0.1851\\ \hline
med & 250 & 50 & 0.0326 & 0.0287 & 0.0286 & 0.0301\\ \hline
std & 250 & 50 & 0.3490 & 0.2091 & 0.2110 & 0.3439\\ \hline\hline
avg & 500 & 100 & 0.1024 & 0.0692 & 0.0790 & 0.1262\\ \hline
med & 500 & 100 & 0.0348 & 0.0335 & 0.0342 & 0.0345\\ \hline
std & 500 & 100 & 0.2184 & 0.1340 & 0.1459 & 0.2396\\ \hline
\end{tabular}
\label{table-diameters}
\vspace*{-1.5mm}
\end{table}
}

\section{Conclusions}

In this paper, we studied the problem of building nonparametric, nonasymptotic, simultaneous confidence regions for band-limited functions. After summarizing a recently developed framework, we proposed two types of improvements:
(1) We tightened the kernel norm upper bound using terms from the uniformly-randomized Hoeffding's inequality and an empirical Bernstein bound. We derived an approximate sample-size threshold, as well, to decide which bound to apply.
(2) We explored majority-voting schemes and showed that even when combining the confidence intervals of each input separately, the resulting intervals still guarantee simultaneous coverage. 
We also demonstrated empirically that aggregation reduces both the variance and the size of the regions.

\bibliographystyle{ieeetr}
\bibliography{references} 

\end{document}